\documentclass[lettersize,journal]{IEEEtran}
\usepackage{amsmath,amsfonts}
\usepackage{algorithmic}
\usepackage{algorithm}
\usepackage{array}
\usepackage[caption=false,font=normalsize,labelfont=sf,textfont=sf]{subfig}
\usepackage{textcomp}
\usepackage{stfloats}
\usepackage{url}
\usepackage{verbatim}
\usepackage{graphicx}
\usepackage{cite}
\usepackage{algorithm}
\usepackage{algorithmic}
\usepackage{booktabs}
\usepackage{multirow}
\usepackage{amsmath}
\usepackage{subfloat}
\usepackage{array}
\usepackage{pifont}
\usepackage{graphicx} 
\usepackage{color}
\usepackage{ragged2e}
\usepackage{cite}
\usepackage[numbers]{natbib}
\begin{document}

\title{Data Quality-aware Mixed-precision Quantization via Hybrid Reinforcement Learning}

\author{Yingchun~Wang,
Song~Guo,~\IEEEmembership{Fellow,~IEEE,}
Jingcai~Guo,~\IEEEmembership{~Memeber,~IEEE,}\\
Weizhan~Zhang,~\IEEEmembership{~Memeber,~IEEE,}
~and~Jie~Zhang,~\IEEEmembership{~Student~Memeber,~IEEE}
\IEEEcompsocitemizethanks{\IEEEcompsocthanksitem Y. Wang is with Department of Computer Science and Technology, Xi'an Jiaotong University, Xi'an 710049, China, and with Department of Computing, The Hong Kong Polytechnic University, Hong Kong SAR., China.\protect\\
E-mail: daenerys.wang@connect.polyu.hk.
\IEEEcompsocthanksitem S. Guo, J. Guo and J. Zhang are with Department of Computing, The Hong Kong Polytechnic University, Hong Kong SAR., China.\protect\\
E-mail: \{song.guo, jc-jingcai.guo, jie-comp.zhang\}@polyu.edu.hk.
\IEEEcompsocthanksitem W. Zhang is with Department of Computer Science and Technology, Xi'an Jiaotong University, Xi'an 710049, China.\protect\\
E-mail: zhangwzh@xjtu.edu.cn.}
\thanks{Manuscript received November 23, 2022.}}

\markboth{Journal of \LaTeX\ Class Files,~Vol.~14, No.~8, August~2021}%
{Shell \MakeLowercase{\textit{et al.}}: A Sample Article Using IEEEtran.cls for IEEE Journals}

\IEEEpubid{0000--0000/00\$00.00~\copyright~2021 IEEE}

\maketitle

\begin{abstract}
Mixed-precision quantization mostly predetermines the model bit-width settings before actual training due to the non-differential bit-width sampling process, obtaining sub-optimal performance. Worse still, the conventional static quality-consistent training setting, i.e., all data is assumed to be of the same quality across training and inference, overlooks data quality changes in real-world applications which may lead to poor robustness of the quantized models. 
In this paper, we propose a novel \textbf{\underline{D}}ata \textbf{\underline{Q}}uality-aware \textbf{\underline{M}}ixed-precision \textbf{\underline{Q}}uantization framework, dubbed DQMQ, to dynamically adapt quantization bit-widths to different data qualities. The adaption is based on a bit-width decision policy that can be learned jointly with the quantization training. Concretely, DQMQ is modeled as a hybrid reinforcement learning (RL) task that combines model-based policy optimization with supervised quantization training. By relaxing the discrete bit-width sampling to a continuous probability distribution that is encoded with few learnable parameters, DQMQ is differentiable and can be directly optimized end-to-end with a hybrid optimization target considering both task performance and quantization benefits. 
Trained on mixed-quality image datasets, DQMQ can implicitly select the most proper bit-width for each layer when facing uneven input qualities.  
Extensive experiments on various benchmark datasets and networks demonstrate the superiority of DQMQ against existing fixed/mixed-precision quantization methods.
\end{abstract}


\section{Introduction}\label{sec:introduction}
\IEEEPARstart{M}{o}del quantization has attracted increasing attention recently as an effective solution that improves the computation and memory efficiency for deep neural networks (DNN), thus enabling the deployment of deep models on resource-constrained edge devices~\cite{wang2022survey,post,HAWQ-v2,deep,dorefa,pact,Lu2019Lip,zhang2018deepvoice}. Specifically, the lower description bit-width of DNN can bring about significant benefits in terms of inference acceleration and model compression rate, for example, Int8/Int4 quantization and binary neural networks~\cite{irnet}. However, it inevitably causes a greater accuracy drop than higher description bit-width models due to unavoidably more information loss. 
To maximally excavate the description redundancy and learn the compressed model for a given training domain, mixed-precision quantization~\cite{HAWQ-v2,hawq,HAWQ-V3,daam,FracBits,post,AutoQ} has been proposed to explore a better balance between task performance and model description length.  
It allows network components, e.g., layers~\cite{haq} or kernels~\cite{AutoQ}, to be quantized with lower bit-widths instead of full precision bit-widths if they have less contribution to model accuracy, wherein, more sensitive components are quantized with relatively higher bit-widths to avoid large accuracy degradation and vice versa. 

Despite the more balanced compression power, the mixed bit-widths can bring about unique challenges for the quantization setting exploration and model training. Concretely, the search space for determining a sufficiently good mixed-precision quantization setting can grow exponentially with increasing number of layers and bit-width choices. 
For example, when a ResNet-152 is quantized layer-wisely with 4 possible bit-widths, the search space can easily reach $4^{152}$ times. Worse still, the inherently discrete and non-differentiable bit-width decision-making (or sampling process) makes it hard to learn the bit-width policy jointly with quantization training. 
To deal with the above challenges, existing state-of-the-art (SOTA) works usually decouple the non-derivable bit-width sampling process from the end-to-end supervised training, i.e., consisting of two mutually independent precision decision-making stage and quantization training stage. 
Some representative methods include state-of-the-art HAWQ series~\cite{HAWQ-v2,hawq,HAWQ-V3}, DMBQ~\cite{daam}, FracBits~\cite{FracBits}, AutoQ~\cite{AutoQ}, etc. 
%

For example, The HAWQ series~\cite{HAWQ-v2,hawq,HAWQ-V3} measure the relative sensitivity of layers via layer-wise hessian information and develop a Pareto frontier based method to make hessian-based bit-width decisions before model quantization. 
%
Differently, DMBQ~\cite{daam} tries to search for an optimal bit-width setting over the distribution space beforehand and quickly selects one setting via this prepared fast lookup table during training. 
In general, previous
These state-of-the-arts make satisfying bit-width decisions under the guidance of hardware feedback~\cite{haq} or gradient-based information~\cite{hawq} before the actual quantization training, as well as the practical inference, while all their bit-width selection and quantization training are done in the given narrow data domain with quality-consistent training setting. \par

\begin{figure}[t]
    \centering
    \includegraphics[width = 0.87\columnwidth]{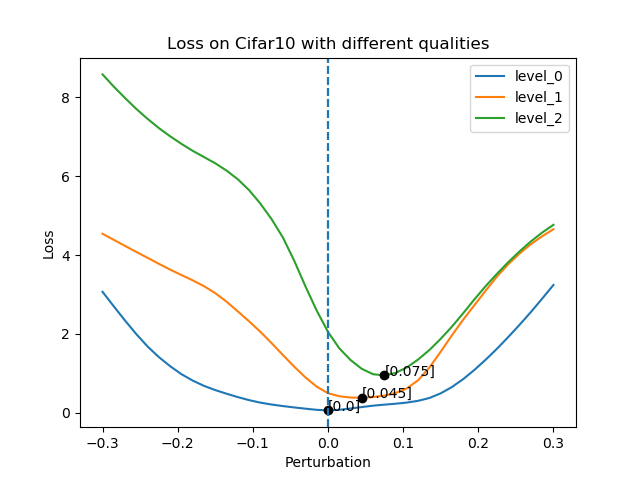}
    \caption{Loss changing of quantization model with different data qualities. `level' denotes the degree of Gaussian blur, i.e., the larger the value, the blurrier the image. }
    \label{losss}
\end{figure}

In real-world application, such decoupled training process can inevitably lead to the data isolation issue between three stages of precision decision-making, quantization training, and practical inference, which can result in sub-optimal performance. 
Besides, the static quality-consistent training setting is not robust against data quality changes in practice, which may further deteriorate the effect of the prior bit-width decision and obtain poor robustness of the quantized models. 
Notably, our observation suggests a non-negligible view that the data quality can have a significant impact on the quantization setting decision due to the changing of relative layer sensitivities to quantization in networks. 
As demonstrated in Fig.~\ref{losss}, we show a demo case of the loss changing on different data qualities, where `level\_0' denotes the quantized model trained on standard CIFAR-10 dataset. The input images are getting blurrier from `level\_0' to `level\_2'. 
To augment images with various qualities, we apply random quantization perturbations to the original quantization model along the direction of the Hessian matrix at the convergence point. 
Assume the data quality does not alter the optimal quantization settings, the minimum task loss should occur when the perturbation is 0. 
However, we get the opposite result. The loss-curve is as expected only if the quality of the test image is the same as the training image. The optimal model description bit-widths can change when the quality of the test image varies, i.e., more specifically, the greater the difference of qualities, the larger the deviation of the model parameters w.r.t the minimum loss point. 
Horizontally, we can also observe that as the image noise increases, the deviation of the quantization convergence point from the original parameters becomes larger. 
In summary, the above findings indicate a huge contradiction between previous quantization works and the need to infer on environment-sensitive devices with unpredictably and dynamically changing input qualities. 

In this paper, our interest is to explore the optimal compressed model by mixed-precision quantization from a new perspective, \textit{i.e., data rather than the model}, and to improve both the generalization and robustness of quantized models. The motivation stems from the fact that, in real-world applications, it is usually impossible to control the source distribution of data, such as camera blur, transmission noise, etc. 
To deal with such issues, we propose a Data Quality-aware Mixed-precision Quantization method, dubbed DQMQ, consisting of two lightweight plug-ins to facilitate data quality-aware bit-width decision-making and one-shot quantization model training, respectively. 
First, the Precision Decision Agent (PDA) plug-in can automatically generate the current optimal layer-wise quantization bit-widths setting during forward propagation. 
Then, the Quantization Auxiliary Computer (QAC) plug-in is used to eliminate the accumulation of quantization biases due to the undetermined precision decision policy. 
Moreover, by feeding mixed-quality samples, the discrete bit-width sampling distribution can be learned jointly with model parameters in a one-shot and continuous scheme. 
%
Being small and light, our plug-ins also hold the following features: 1) be deployed in the training phase and can be removed flexibly during other phases; and 2) can be easily applied to most mainstream neural networks. 
%
%


Our contributions are summarized as follows:

\begin{itemize}
\item We first indicate that the data quality can significantly affect the layer-wise quantization sensitivity, and upon which, we propose a novel data quality-aware mixed-precision quantization method to address such issues.

\item We incorporate the previously decoupled bit-width precision decision-making model parameter training into a one-shot scheme, which significantly improves the efficiency and resolves the data isolation. Moreover, we also provide a theoretical derivation to prove its feasibility. 

\item Experiments on various datasets and mainstream networks demonstrate the effectiveness of our method. In particular, our DQMQ outperforms the SOTA HAWQ-v2 by 12.7$\%$ in MAP and 7.30$\times$ in model compression rate, respectively. 
\end{itemize}

\section{Related Works}
\subsection{Mixed-Precision Quantization} 
An intractable issue in conventional fixed-precision model quantization is that high bit-width quantization can ensure high precision, while the memory footprint and calculation cost are also sufficient large. On the contrary, the memory usage and calculation cost can be smaller with lower quantization precision, while this unavoidably leads to significant quantization accuracy degradation. This essential dilemma makes quantization under fixed bit-width can hardly achieve a fine-grained trade-off between accuracy and FLOPs. 
To fully exploit the model redundancy and obtain more balanced compression power, mixed-precision is proposed. 
It achieves further efficient compression of the model and brings better quantization benefits due to precision flexibility~\cite{DBLP:journals/corr/abs-1812-00090, post}. For example, DMBQ~\cite{daam} searches the optimal quantization scheme over the distribution space beforehand and selects the quantization scheme during training using a fast lookup table-based strategy. Wang \textit{et al.}~\cite{haq} estimate layer-wise resource cost by simulators, and combine the hardware indicators to constrain the bit-widths search for each layer. Later, offline bit-widths decision-making based on the layer-wise Hessian Spectrum is enabled in HAWQ serious~\citep{hawq, HAWQ-v2, HAWQ-V3}.
These studies have produced outstanding results when compared to single-precision quantization. 
However, the majority of them decouple the actual quantization training and quantization bitwidth selection due to the discrete mixed bitwidth selection procedure. 
In other words, even when the quality and distribution of the input data are expected to change throughout quantization training, they consistently believe the pre-selected bit width to be the best option. 
And our experimental results in the previous section have proved that this is wrong.

\subsection{Quantization Aware Training} 
Quantization Aware Training (QAT) fits feature inputs to quantized parameters that are as low as possible in bit-width while minimizing the loss in optimization target jointly in the training process. Specifically, a typical QAT pipeline simulates a quantization procedure by introducing fake quantization operations to model weights and activations during the forward propagation, while the gradient update is still performed on vanilla float parameters during the backpropagation as if the quantification never happened \cite{Jacob_2018_CVPR}. However, because of the inherently discrete and non-derivative quantization operations, QAT cannot calculate accurate gradients like a standard smooth neural network in backpropagation. A common practice is to use a gradient straight-through estimator (STE)~\cite{ste1}, which fits the parameters of a discontinuous function to an approximately continuous function and ignores the impact of quantization. Esser \textit{et al.}~\cite{esser} introduces a novel means to estimate and scale the task loss gradient for each weight and activation and enable it to be learned in conjunction with other network parameters. Although these methods under QAT schemes achieve excellent performance, most of them require great resource cost and significant time on re-training and hyper-parameter tuning \cite{updown}. Our work removes these redundancies through one-shot training and a lightweight quantizer.

\subsection{Post-training Quantization}  
Post-training Quantization (PTA) quantizes neural networks during practical deployment after the standard model training stage. In the quantization process, it selects appropriate quantization and/or calibration operations for the pre-trained network to minimize quantization loss. Without re-training, PTA is simpler to use and allows for quantization with limited data conditions. For example, many PTA works need only a small calibration set: Fang~\textit{et al.}~\cite{fang2020post} proposed Piecewise Linear Quantization (PWLQ) to solve the great accuracy drop in uniform quantization with lower bit-width. It finds optimal segmentation breakpoints to obtain lower quantization errors and only needs a small amount of calibration data. 
Furthermore, there are even some PTA works that are data-free during the quantization process. DFQ \cite{1} does not require re-training, nor does it require calibration data to determine quantitative parameters. It uses the mathematical properties of the activation function ReLU to balance the data range of each channel of the adjacent two layers of weights to compensate for the quantization errors. However, 
although PTQ is easier to use, when deployed on moving edge models with dynamically changing data quality, the isolation of data information at training time and at quantization time may lead to an obvious degradation of the generalization of quantized models.

\section{Optimization Theory} 
\subsection{Quantization Preliminaries}
Given a dataset with $N$ labeled samples as $X = \left \{ x_i \right \}_{i=1}^{N}$, $Y = \left \{ y_i \right \}_{i=1}^{N} $.
For a CNN with $L$ convolution layers, we define $\Theta$ as the learnable parameters set, and $\omega_l\in \Theta$ as the vanilla full precision weight parameters of layer $l$. 
A typical quantization-aware training CNN structure can be described as an intertwined pipeline including quantization $\rightarrow$ convolution $\rightarrow$ dequantization. 
\begin{figure*}[htp]
    \centering
    \includegraphics[width = 0.97\textwidth]{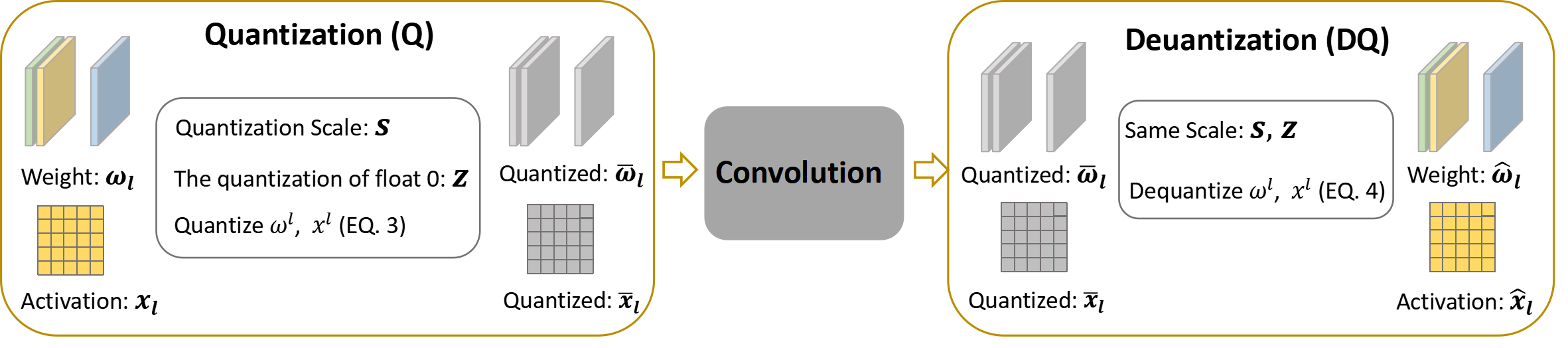}
    \caption{The quantization of convolution architecture. }
    \label{quantization}
\end{figure*}
As shown in Fig.~\ref{quantization}, the left tangle shows the process of the quantization of the convolutional input, wherein the $\omega_l$ and the $x_l$ are the full precision model weight and activations respectively. $s$ is the minimum scale that can be represented after fixed-point quantization, and $z$ represents the quantized fixed-point value corresponding to the full-precision 0 floating-point value (For brevity, the following Eq. (\ref{1}), Eq. (\ref{1}), and Eq. (\ref{3}) only show the computing of $\omega$, and the activation calculation is the same):
\begin{equation}
\begin{aligned}
    s=\frac{\omega^l_{max} - \omega^l_{min}}{\bar{\omega}^l_{max}-\bar{\omega}^l_{min}},~~~z=\bar{\omega}^l_{max}-  \frac{\omega^l_{max}}{s}.
\end{aligned}
\label{1}
\end{equation}
The quantization formula from floating point to fixed point is as follows:
\begin{equation}
\begin{aligned}
    \bar{\omega}_l = \frac{\omega_l}{s}+z.
\end{aligned}
\label{2}
\end{equation}
Note that the weight quantization is completed before inference, and the activation quantization needs to be performed during inference. After $Q$ (the quantization operation), the $\bar{\omega}_l$ and $\bar{x}_l$ represent the quantized values for actual convolution calculation. \par
The right tangle represents the dequantization operation, and $\hat{\omega}_l$ and $\hat{x}_l$ represent the dequantized weight and activations respectively. 
The dequantization formula from fixed point to floating point is as follows:
\begin{equation}
\begin{aligned}
    \hat{\omega}_l=(\bar{\omega}_l-z)*s.
\end{aligned}
\label{3}
\end{equation}
Previous works only aim to train model parameters with statically predetermined bit-width settings, for what most of them perform indefinite pseudo-quantization attempts depending on the quantization ability of the next operator.
Differently, DQMQ can not only to train model parameters to shorter bit-widths for the current data domain, but also learn to sense the changes in data quality and dynamically make optimal bit-width decisions to cope with uncontrollable test distribution shifts in the real world ( such as additive blur, transmission noise or compression aberration).
Thus, in our work, the $DQ$ (dequantization) is inserted after each convolution layer for the data aware mixed precision pseudo quantization.
not only produces full precision inputs for the subsequent layer, but also need to return the convolution layer to an unquantized state in preparation for bit-width sampling in the next iteration (preventing the accumulation of quantization errors).

1) Symbols for PDA: For any layer $l$, PDA inserted before $l$, named $Agent_{l}$, is the RL agent for precision-decision making. Define $h_l$ as the hessian trace of layer $l$ indicating the sensitivity of layer $l$ and $x_l$ as the originally unbiased full precision input features for layer $l$. Thus, for layer $l$, input $x_l$ and $h_l$, PDA is trained to give the optimal quantization precision $a_l$. \\

2) Symbols for QAC: The quantizer and dequantizer in QAC are defined as $Q$ and $DQ$, respectively.
Thus, we can use parameters with bars for the ones quantized by $Q$,
and parameters with hats for full-precision ones dequantized by $DQ$. For brevity, we still use $x_l$ and $\omega_l$ to represent the input features and weights of $l_{th}$ layer in the context).  
\par
\subsection{Minimum Description Bit-width Sampling}
Model quantization follows the principle of minimum description length in information theory, which is defined as conveying the sum of complexity and misfit of models in the least number of bits. Thus, we define the skeleton of the optimization target for DQMQ as the hybrid minimization of both model errors and memory complexity. 
More specifically, the model errors is targeted as supervised learning and the model size reduction can be defined as the reward in policy optimization problem. High model bit-widths obtains better task performance with low memory reward,  while ultra low bit-width lead to
significant accuracy degradation with more memory benefit. 
As a mixed precision method, DQMQ samples the optimal precision for each convolution layer in the confrontation between task accuracy and memory gain. Bit-width sampling is defined as an inherently discrete and therefore non-differentiable decision process based on several possible bit-widths like Int2, Int4, Int8, etc. It is encoded as a probability distribution function mapping feature input and layer quantization sensitivity to the layer-wise quantization bit-width.
Note that the bit-width sampling should be done before the parameters quantization and the convolution forward propagation. Thus, in our work, the $Q$ (quantization) is inserted before each convolution layer to make data-aware bit-width decisions.
We will provide the details of the optimization formulation of DQMQ in the next subsection.
\subsection{Optimization Objective}
Define $\textbf{a}=\{a_1, a_2, ... , a_L\}$ as the set of quantization precision decisions and $\textbf{h}=\{h_1, h_2, ... , h_L\}$ as the set of relative quantization sensitivities for all layers. According to the above discussions, the optimization target of DQMQ can be defined as:
\begin{equation}
\begin{aligned}
\min J(\Theta) &= \min E_{\textbf{a}|X, \textbf{h}} L_{\Theta}(X,\textbf{a},\textbf{h})\\
&= \min E_{\textbf{a}|X, \textbf{h}}
\left[\mathcal{L}(\hat{Y}(\bar{X}, \bar{\Theta}, \textbf{a}), Y) - \alpha\sum_{l=1}^{L}R_l\right] \\
&= \min P(\textbf{a}|X,\textbf{h}) \left(\mathcal{L}_{\Theta} - \alpha\sum_{l=1}^{L}R_l\right),
\label{4}
\end{aligned}
\end{equation}
where $\mathcal{L}_{\Theta}=\mathcal{L}(\hat{Y}(\bar{X}, \bar{\Theta}, a), Y)$, $\Theta$ is the learnable parameters set and $\mathcal{L}$ is the loss function. $\hat{Y}(\bar{X}, \bar{\Theta})$ means that model outputs are actually dequantized values of the convolution results on quantized weights and activations.
$R_{l}$ is defined as the benefit brought by $a_l$-bits quantization of layer $l$, which is defined as model size reduction in our work. Finally, $\alpha$ is the weighting factor that allows us to trade-off the opposite optimization target of minimizing empirical risk loss and maximizing quantization reward. \par
The learnable key elements (policy $\pi$ and future cumulative reward $r_l$) for $Agent_l$ are defined as follows:
\begin{equation}
\begin{aligned}
\mathrm{Bitwidth~Policy}:\\
&\pi(x_l, h_l) = P_{\Theta}(a_l|x_l, h_l),\\
\mathrm{Future~Reward}:\\
&r_l = \alpha \sum_{i=l}^{L}R_i - \mathcal{L}_{\Theta}.
\label{5}
\end{aligned}
\end{equation}
To jointly optimize Eq. (\ref{4}) and Eq. (\ref{5}) by gradient descent, the gradients of Eq. (\ref{4}) can be derived as:
\begin{equation}
\begin{aligned}
&\frac{\partial J(\Theta)}{\partial \Theta} = \bigtriangledown_\Theta P_\Theta(\textbf{a}|X, \textbf{h}) L_{\Theta}(X,\textbf{a}, \textbf{h}) \\
&= P_\Theta(\textbf{a}|X, \textbf{h}) \bigtriangledown_\Theta   \mathcal{L}
+ L_{\Theta}(X,\textbf{a}, \textbf{h}) \bigtriangledown_\Theta \prod_{l=1}^{L} P_{\Theta}(a_l|x_l, h_l) \\
&=E_{\textbf{a}|X, \textbf{h}}\left\{\bigtriangledown_\Theta \mathcal{L}_\Theta + {\bigtriangledown_\Theta \log P_\Theta(\textbf{a}|X, \textbf{h}) L_{\Theta}(X, \textbf{a}, \textbf{h})}\right\}.
\label{6}
\end{aligned}
\end{equation}

\section{Methodology}\label{sec3}
In this section, we describe the proposed DQMQ in details from three main aspects including an architecture overview, the components' details and the deployment workflows in the real-world. 
Specifically, DQMQ is a one-shot quantization system which includes two intertwined pipelines: bit-widths decision-making and model parameters updating. DQMQ is modeled as a hybrid reinforcement learning task via two plug-ins: Precision Decision Agent (PDA) and Quantization Auxiliary Computer (QAC). Being small and general, they can be inserted into most of the popular CNNs. We introduce the overview of the architecture of DQMQ in subsection. \ref{section:4.1} . Then, we detail the details of both plug-ins to explain why DQMQ has to be carefully designed like this in subsection. \ref{section:4.2} and subsection. \ref{section:4.3}. Finally, we show the deployment workflow in practice in subsection. \ref{section:4.4}.

\subsection{DQMQ: An Overview}{\label{section:4.1}}
To jointly learn dynamical data quality-aware bit-widths and model quantization parameters in one-shot training, it is necessary to make the inherently discrete and therefore non-differentiable decision process learnable. 
Moreover, during the training stage of DQMQ, the bit-width decisions across iterations are different due to the dynamically changing input quality. Suppose that the bit-width decision module in this round gives the answer of Int4, and in the next round, the unconverged bit-width decision module gives another decision of Int8 bits. If layer-wise dequantization is not performed in each round as in previous mixed-precision SOTAs, the training error of model parameters brought by each quantization attempt will gradually accumulate and seriously affect the training accuracy. Therefore, DQMQ should take into account how to isolate different quantization attempts on dynamically changing data so that the training in each round would not affect each other.\par
First, in order to cope with the dynamically changing data quality and make the corresponding optimal quantization bit-width decision, we first carefully encode PDA to simulate the discrete probability distribution over the bit-widths candidates. We solve the non-differentiable training of PDA by regarding it as an agent in RL paradigm, and learn the policy of making optimal bit-width decisions in a policy optimization framework. 
After being trained, PDA generates the most likely bit-width action for each layer according to the current input features and layer quantization sensitivities. The details about PDA will be discussed in Sect.~\textit{PDA: Bit-width Decision-making}.

Afterward, the quantizer and the dequantizer of QAC will work together to make fake quantization attempts to compute the current quantization reward. Receiving the most ideal bit-widths from PDA, the quantizer in QAC outputs the quantized model weights and activations with ideal bit-width, which would be the actual convolutional operators to generate feature maps for the next layer.
The dequantization operation just follows the convolution operation. It not only makes sure that the next layer receives the correct full-precision activations to eliminate the quantization error accumulation layer by layer in one forward pass, but also stores the full-precision weights of the network to avoid the accumulation of quantization parameters updating biases due to different bit-widths actions from two nearby iterations. The algorithm details and proof of QAC will be shown in Sect.~\textit{QAC: One-shot Training}. 
\begin{figure*}[t]
\centering
\includegraphics[width=0.92\textwidth]{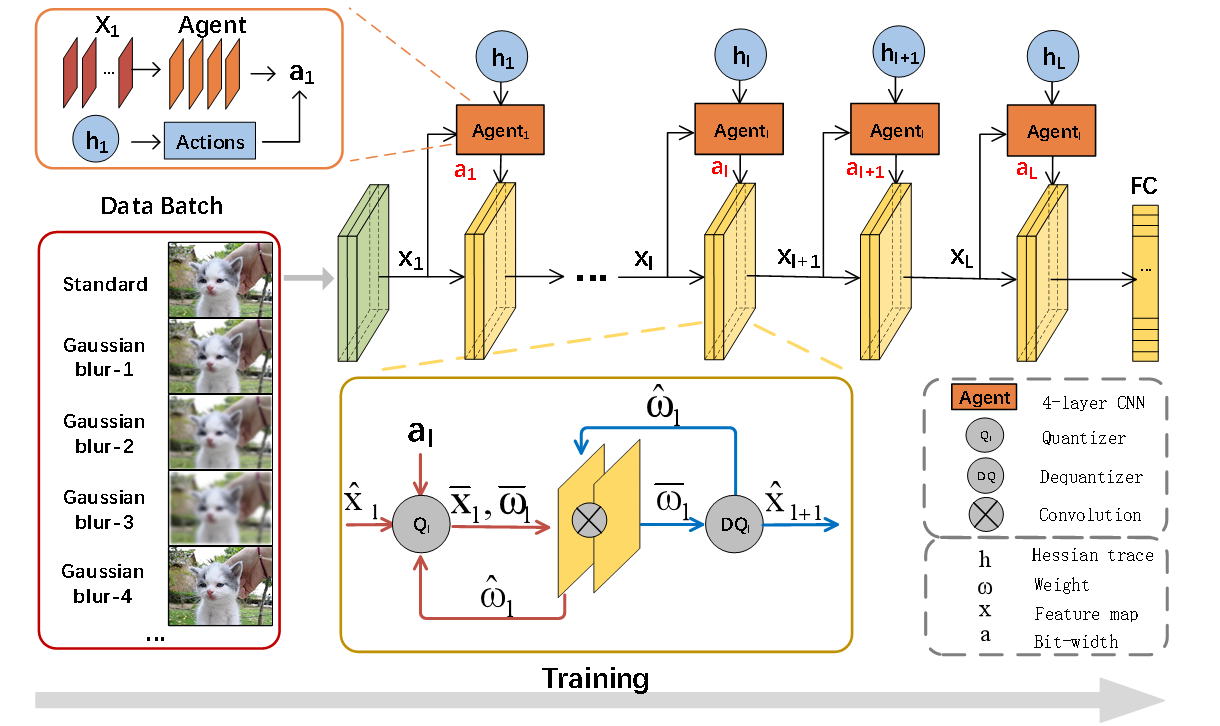}
\caption{Overall architecture design of DQMQ.}
\label{fig2}
\end{figure*}

More specifically, Fig.~\ref{fig2} shows how to integrate the proposed plug-ins with the existing training pipeline.
For a specific layer (e.g., $l$), the orange rectangle $Agent_{l}$ corresponds the above-mentioned PDA module. Encoded by a tiny four-layers CNN, PDA maps the input $x_{l}$ and sensitivity indicator $h_l$ to a bit-width decision $a_l$, of which $a_l$ is actually a candidate in the action pool of layer $l$ determined by $h_l$. 
More specifically, for quantization-sensitive layers with large $h_l$ values, the drop in model accuracy due to quantization disturbance is even greater. Therefore, the bit-width values in the candidate pools of these layers tend to be relatively large, trading a smaller loss of memory gain for a smaller loss of model accuracy. Conversely, for quantization-insensitive layers with small values of h, quantization with lower bits does not bring about a significant change in model accuracy, which means that these layers can be quantized with ultra-low bits. Quantization-insensitive layers will be the main source of DQMQ's quantized memory gains.
%
The yellow cube in Fig.~\ref{fig2} represents the convolutional layer with the added QAC module. The two grey circles represent the quantizer and the dequantizer, which together make up the QAC. The lower yellow box in Fig.~\ref{fig2} shows the workflow of QAC. 
Specifically, the quantizer $Q_l$ in QAC of layer $l$ quantizes the full-precision weights $\omega_{l}$ and activations $x_{l}$ of this layer to $\bar{\omega_{l}}$ and $\bar{x_{l}}$ with $a_l$-bits. As shown by the training rectangle in Fig.~\ref{fig2}, convolution operation is conducted on the quantized operators $\bar{\omega_{l}}$ and $\bar{\omega_{l}}$ to simulate the effect of $a_l$-bits quantization on final model performance. After that, the dequantizer $DQ_{l}$ inserted behind $l$ converts the quantized weight $\bar{\omega_{l}}$ and convolution results back to full-precision weights $\hat{\omega_l}$ and activations $\hat{x_{l+1}}$. 

\subsection{PDA: Bit-width Decision-making}{\label{section:4.2}}
During quantization, the most likely bit-widths are taken from the probability distributions encoded by PDA. It is a four-layer CNN that maps the feature input to a four-class decision. 
To handle this discrete and therefore non-differentiable decision process, we reference the training method used in SkipNet~\cite{skip}. Regarding the bit-width decision-making module as the \emph{$agent$}, the backbone neural network as \emph{environment} giving feedback, the layer-wise bit-width decision as the \emph{action}, the memory reduction $\sum_{l=1}^N{R_l}$ in Eq. (\ref{1}) as \emph{rewards}. We define the cumulative future reward in Eq. (\ref{3}) and learn the policy defined in Eq. (\ref{2}). \par
To learn this RL based bit-width policy synchronously with the normal supervised learning, we use a hybrid learning algorithm ~\cite{skip} which combines empirical risk minimization and cumulative reward maximization. 
Specifically, in the first soft training stage, the output of the PDA is relaxed to continuous probability outputs over different quantization bit-width decisions to train fake-quantized model end-to-end by gradient descent to reduce task loss. Then, to reduce the approximation errors of the optimal quantization setting caused by relaxation, we perform hard training in the second stage to further optimize PDA by RL with respect to model quantization benefit maximization on discrete bit-width decision actions. In more detail, the bit-width decision-making module PDA is modeled as a \textbf{$agent$} in DQMQ, with the backbone neural network as \textbf{environment} giving feedback, the layer-wise bit-width decision as \textbf{action}, the memory reduction $\sum_{l=1}^{L}{R_l}$ in Eq. (\ref{4}) as \textbf{rewards}, and also the \textbf{cumulative future reward} and \textbf{policy} in Eq. (\ref{5}).\par
To address the huge RL search space of previous RL quantization works, we assign different action pools to different layers based on the hessian trace information, which has been proved as an effective quantization sensitivity indicator~\cite{HAWQ-v2}. Specifically, there may be only ultra-low bit-width candidates like Int2 or Int4 in the action pools of layers with lower quantization sensitivity. For more sensitive layers, their bit-width candidates are relatively higher like Int8 or Int16. In our work, there are three kinds of action pools $[0,1,2]$, $[2,4,8]$ and $[8,16,32]$.\par

\subsection{QAC: One-shot Training}{\label{section:4.3}}
Most state-of-the-art approaches for mixed-precision quantization start from a decoupled bit-widths search stage. They find the optimal quantization bit-width setting before actual quantization training instead of dynamically learning it. However, invariant data prior during quantization training makes the learned quantized model only a local optimum within this data quality region and loses generalization ability on data of different quality.
This is not consistent with variational data qualities in real-world and may cause serious problems when the lightweight quantized models are practically deployed in edge applications.  
To solve the data isolation problem, we incorporate the previously decoupled bit-width setting search and model parameters training into one-shot training. This creates challenges as follows. \textbf{Between layers:} Different layers have a high probability of being quantized to different bit-widths. As the inputs of the next layer, the quantized activations of the current layer would affect the joint convolution and quantization of the next layer. \textbf{Between iterations:} The optimal bit-widths settings in each forward propagation would be different due to the undetermined precision decision policy. This brings different optimization precision in each iteration, and further results in the accumulation of optimization biases, which may lead to training crushing.\par
To facilitate one-shot training, we creatively propose a method named QAC that could optimize the backbone network with different bit-widths without biases accumulation. QAC is composed by the quantizer $Q$ and a the dequantizer $DQ$. The detailed plug-ins design and the optimization process and  are shown in Algorithm.\ref{alg1:Framwork} and \ref{alg2:Framwork}. 
\begin{algorithm}[tb] 
\caption{ The forward process of training stage of DQMQ. } 
\label{alg1:Framwork} 
\begin{algorithmic}[1] 
\REQUIRE ~~\\ 
The hessian trace of the $l_{th}$ layer: $h_l$ ;\\
The number of layers to be quantized: $L$;\\
The weights and inputs of the $l_{th}$ layer: $\omega_l$, $x_l$;\\
Parameters with bars are quantized rounded integer values by $Q$; \\
Parameters with hats are dequantized full precision values by $DQ$;\\
The PDA, quantizer and dequantizer of QAC: $Agent_{l}$, $Q_{l}(·)$ and $DQ_{l}(·)$;\\
The policy, reward and action of RL: $\pi_l$, $r_l$, and $a_l$;\\
The classifier layer: $fc$;\\ 
Loss function: $f(·)$;
\ENSURE ~~\\ 
The trained DQMQ;

\FOR{$l=1$ to $L$}
\STATE $a_l = \pi_l(h_l, x_l)$ ;
\STATE $\bar{x}_l = Q_l(x_l,a_l)$, $\bar{\omega}_{l} = Q_l(\omega_l,a_l)$;
\STATE $\bar{x}_{l+1} = \bar{x}_l \otimes \bar{\omega}_{l}$;
\STATE $\hat{x}_{l+1} = DQ_{l}(\bar{x}_{l+1}), \hat{\omega}_l = DQ_{l}(\bar{\omega}_{l})$;
\ENDFOR
\STATE $\hat{y}_{pre} = DQ_{L}(fc(\hat{x}_L))$;
\STATE Compute the optimization target defined in Eq. (\ref{1})

\end{algorithmic}
\end{algorithm}

\begin{algorithm}[htb] 
\caption{The backward process of DQMQ training.} 
\label{alg2:Framwork} 
\begin{algorithmic}[1] 
\REQUIRE ~~\\ 
$net(x,\omega)$ represents the backbone of DQMQ with input $x$ and convolution weight $\omega$;\\
$\otimes$ represents the convolution operation;\\
$Q(x) = Round(\frac{x}{s_1}+z)$, $Q(w) = Round(\frac{\omega}{s_2}+z)$;\\
$DQ(x) = (x-z)*s_1$, $DQ(w) = (\omega-z)*s_2$, \\
$DQ(y) = y*s_1*s_2$;\\
where $s_1=\frac{2^\textbf{a}-1}{\max_x-\min_x}$,
$s_2=\frac{2^\textbf{a}-1}{\max_{\omega}-\min_{\omega}}$, and $z=0$ in our work;\\
\ENSURE ~~\\ 
The trained DQMQ;

\STATE Algorithm 1: $\hat{y}_{pre}(fc(\hat{x}_L))$ ;

\STATE $\hat{y}_{pre} = DQ(fc(\hat{x}_L))
= fc(net(\hat{x}\otimes\hat{\omega})*s_1*s_2 
=  fc(net(\frac{x}{s_1}),\frac{\omega}{s_2}))*s_1*s_2
= \frac{1}{s_11}\frac{1}{s_2}fc(net(x,\omega))*s_1*s_2 
= fc(net(x,\omega))$;

\STATE $\Delta \omega=\frac{\delta T}{\delta y}*\frac{\delta y}{\delta \omega}$;\\
\STATE This indicates the variables being optimized in training stage are the full-precision ones, which makes the quantization decisions between multi-iterations to be independent.\\
\end{algorithmic}

\end{algorithm}

\subsection{Practical Deployment}{\label{section:4.4}}
\begin{figure*}[t]
\centering
\includegraphics[width=0.90\textwidth]{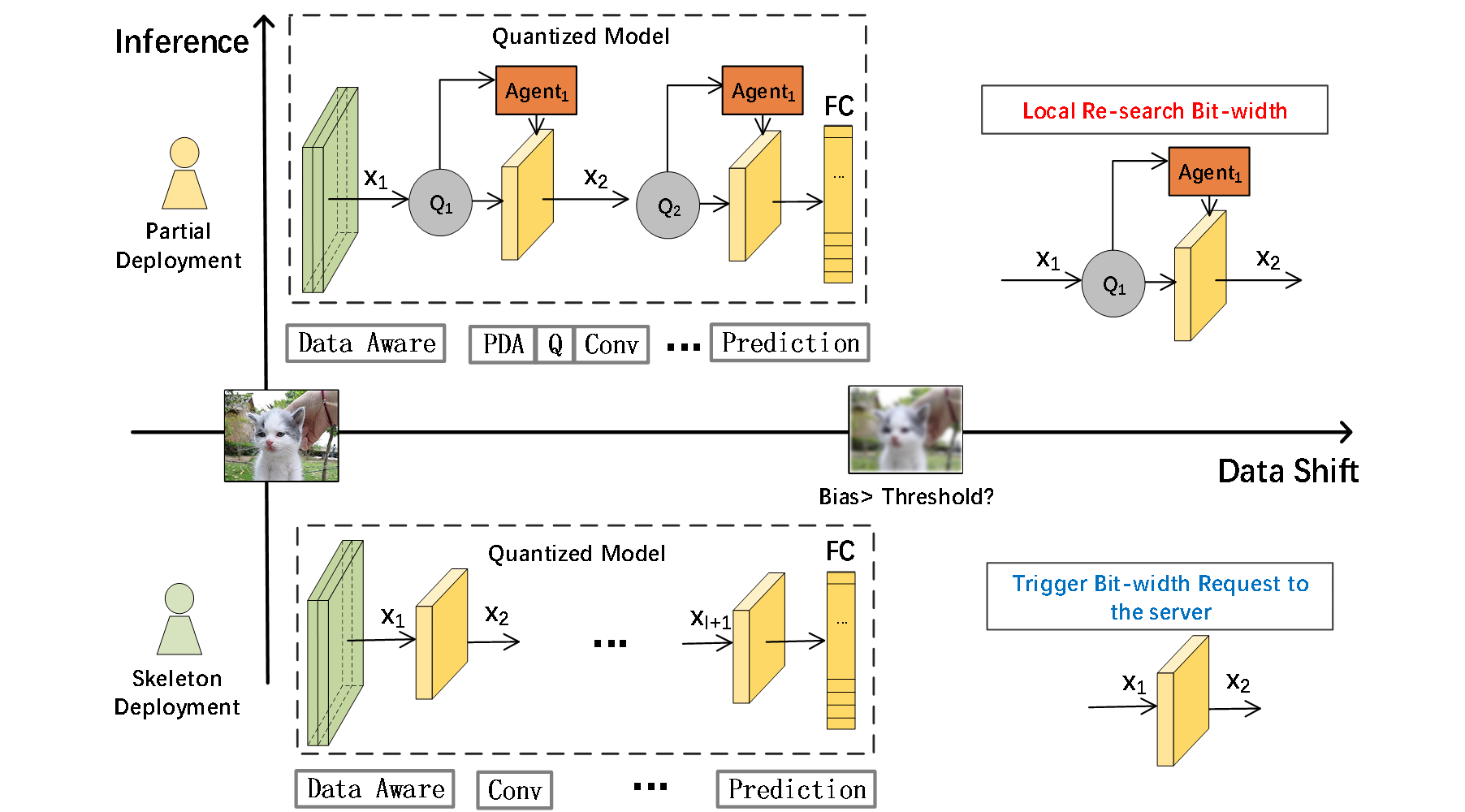}
\caption{Coping mechanisms for partially deployed and skeleton-deployed DQMQs when encountering data quality excursions exceeding a threshold.}
\label{fig4}
\end{figure*}
Despite the limited resources, such as storage, computing units, power, network, etc., the booming embedded devices at the edge provide a vast platform for the deployment of intelligent applications around humans. Model quantization aims to compress deep models into lightweight intelligent applications, reducing the memory and computational complexity of intelligent applications so that they can be deployed on resource-constrained embedded devices. 
But are these lightweight predictive architectures robust enough for real-world inference? Discarding part of the network structure has led to a more or less decrease in the accuracy of these lightweight models. Another potential risk stems from the fact that lightweight intelligent models are faced with real-world inference tasks. , the source distribution of the input data is beyond control. For example, camera blur, transmission noise, image compression, viewpoint changes, etc. can cause the test distribution to shift relative to the static training data. As analyzed in the introduction, dynamically changing test data quality leads to different optimal quantization bit-width settings, despite the same model and the same data class distribution.\par
DQMQ is the first work to quantify the shift in the quality distribution of test data. We have analyzed how DQMQ trains the model at specific times to obtain the perception of dynamic quality in the above subsections.
Although DQMQ adds an additional quantization auxiliary module to the skeleton of a deep neural network, this does not increase the computational or memory overhead of the quantized model he generates during inference. On the contrary, due to data quality-aware quantization, it can maximize model redundancy according to the current data distribution to generate deep quantized models with minimal description length while maintaining task accuracy. Note that the extra quantization auxiliary module is very lightweight (a four-layer fully connected network, two multiplication and division operations and two addition and subtraction operations) compared to the bulky backbone neural network. Also, these quantization-assisted models are not necessarily deployed on actual mobile edge devices. DQMQ will dynamically decide to deploy the entire quantitative model (including PDA and QAC), partial quantitative model (including only PDA and part of QAC), or only the quantized model backbone (excluding any component). We will detail the workflow of DQMQ in practical deployment in this subsection to illustrate the wide applicability of DQMQ in the real world.\par
As shown in Fig.~\ref{fig4}, the deployment of DQMQ can be divided into three situations: \textbf{full deployment}, \textbf{partial deployment} and \textbf{skeleton deployment} according to the resource situation of edge mobile devices. Full deployment refers to deploying the entire DQMQ framework to the edge when the resources of edge mobile devices are sufficient to obtain all the capabilities of DQMQ, including quality-aware bit-width selection, model quantization, and model retraining. Partial deployment is defined as choosing only the data quality perception module (e.g. computing disparity), bit-width decision module (PDA) and quantization module (quantizer in QAC) with the backbone of the quantization model when the resources of the mobile device are not sufficient. They are deployed on edge devices together, so that edge devices can obtain data-aware bit-width selection and model requantization functions. Backbone deployment refers to deploying only the quality perception module and the quantized neural network backbone to the edge when the resources of edge mobile devices are extremely scarce. At this time, the data-aware bit-width selection is jointly completed by the edge side and the server side. When the edge side detects that the data quality has a large deviation (such as day and night, changing the environment), the edge side initiates the server to re-select the bit-width And quantify model application and report current data quality. The server will reselect the quantization bit-width according to the current data quality, and return the new quantization model parameters to the edge device.

To be specific, for data-sensitive mobile applications with relatively adequate memory, such as laptop, workstation and smart phones, etc., PDA and the quantizer of QAC mentioned before are deployed with the backbone of the neural network to make timely adjustments in the face of changing data quality, and it follows the above-mentioned workflow.
For extremely resource-constrained devices like smart watches and Google glasses, all plug-ins would be removed in actual deployment to obtain the most lightweight DQMQ. The necessary adjustments of quantization models when facing unacceptable data quality changes (i.e., image pixel error exceeds a predefined threshold.) will be done by triggering a command to request a new model from the server. The details are as follows:
\ding{182} By setting a very small data quality-aware module at the forefront of the deployed backbone, an unacceptable changing of inputs quality would be detected. 
In details, the data quality-aware module is a resolution-based image-quality detector. Once the resolution deviation exceeds the threshold, for example, a change of shooting scene, the model-request command would be triggered instantly.
\ding{183} Edge devices send the request together with the quality of inputs which triggered the model-request to the edge server.
\ding{184} Receiving a specific quality, the most likely bit-width setting is made through the trained PDA and the quantization is done jointly.
\ding{185} The new lightweight quantization model is sent to the devices.

\section{Experiments}
We compare the proposed DQMQ with a range of state-of-the-art (SOTA) quantization works and evaluate them on three benchmarks: ImageNet, CIFAR-10, and SVHN. We construct the DQMQ of ResNet-18 models by sandwiching layers with quantization and dequantization modules and inserting a decision CNN on each residual block. In Sec. 4.1, we decipher the essence of layer sensitivity to quantization and study its correlation with data quality. In Sec. 4.2, we evaluate the performance of DQMQ and compare it with popular approaches, including mixed precision quantization works DQ, AutoQ, HAWQ~\cite{dq,AutoQ,hawq}, and fixed precision quantization PACT, LQNet, DNAS, US, SAT, TQT networks~\cite{pact,lqnet,dnas,us,sat,tqt}. We also compare our approach with a full precision network to demonstrate the effectiveness of the data quality-aware decision policy.

\subsection{Benchmark}
\label{section:4.1}

\subsubsection{Dataset}
\begin{figure}[t] 
	\centering  
	\subfloat[Original image.]{
		\includegraphics[width=0.4\linewidth]{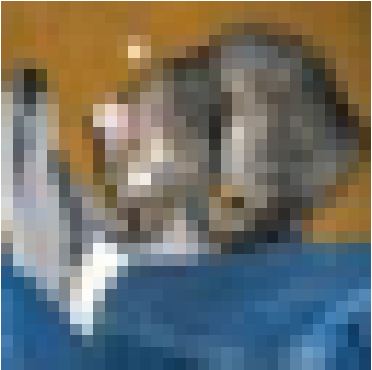}}
	\subfloat[Edge-sharpened image.]{
		\includegraphics[width=0.4\linewidth]{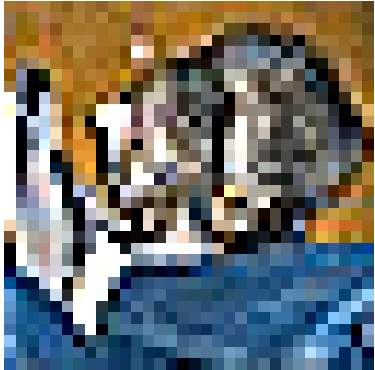}}\\
	\subfloat[Image after Gaussian blur with radius 1.]{
		\includegraphics[width=0.4\columnwidth]{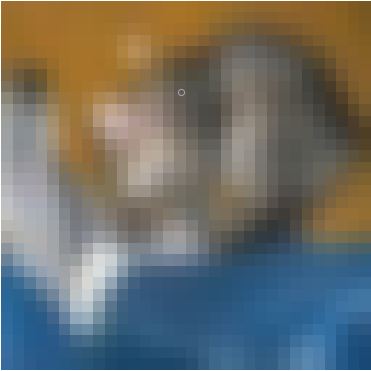}}
	\subfloat[Image after Gaussian blur with radius 2.]{
		\includegraphics[width=0.4\linewidth]{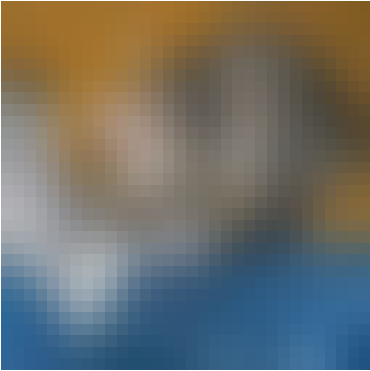}}
	\caption{CIFAR-10 images of different qualities.}
	\label{fig5}
\end{figure}

Extensive experiments are conducted on three popular datasets for model quantization including ImageNet, CIFAR-10, and SVHN. The diversity of data patterns more effectively reflects the generality of our proposed method.
For ImageNet, it has greater scale and diversity, thus providing more reliable evaluations of the proposed vision methods. Specifically, ImageNet is currently the largest database for image recognition in the world. Over 14 million images were manually annotated by ImageNet to indicate objects in the picture. In at least one million images, bounding boxes are also provided. ImageNet contains more than 20,000 categories. In this work, we use the lightweight ISLVRC2012 dataset with 1000 categories. Each class contains hundreds of images. Among them, the training subset contains more than 1.2 million natural images, and the validation subset contains 50,000 images. We process the image size as 224 $\times$ 224.
CIFAR-10 is a color image dataset that approximates ubiquitous objects. It contains a total of 10 categories of RGB color pictures and the size of each image is 32 × 32 There are 6000 images in each category, and there are a total of 50000 training images and 10000 testing images in the dataset. Note that we have enhanced the image quality of CIFAR-10. Specifically, to simulate the dynamically changing data quality on mobile devices, we first processed CIFAR-10 with a low-pass filter with Gaussian blur radius of 1, 3, and 5 to represent varying degrees of low-quality input.
After that, we perform image sharpening by enhancing the edge position based on second-order differential to represent high-quality input that is easier to identify. Fig. \ref{fig5} provides an intuitive illustration of the above processing results.
We use `level 1' to `level 5' to name the five different picture qualities, the bigger the number, the blurrier the image. `level 1' denotes the sharpening image and `level 2' denotes the raw data. We select 10000 images from each of the above five scales respectively and ensure each selection is done with uniform data label distribution. These images with 5 different qualities are mixed together to form a new dataset named `CIFAR-10 (Mixed)'.
The Street View House Numbers (SVHN) dataset is a collection of house numbers taken from Google Street View imagery and contains more than 600,000 labeled data of digit images for identifying alphanumeric characters in natural scene images. The dataset contains ten categories, of which the training set contains 73,257 images and the test set contains 26,032 images. Each image size is 32 $\times$ 32. Note that SVHN undergoes the same mixed-quality treatment as CIFAR-10.

\subsubsection{Models} 
Based on multiple datasets with diverse image patterns and uneven data qualities, we explore the effectiveness of the proposed DQMQ on ResNet-18 for it is the most commonly used network by state-of-the-art quantization works. 
To be specific, ResNet-18 has 4 residual blocks and 2 convolution layers in each block. 
Note that the skeleton neural network is improved based on the design concept of DQMQ. To be specific, each convolutional layer is sandwiched by a data-aware module PDA and a quantization auxiliary module QAC.
The detailed implementation is as follows. Before each convolution layer, a tiny four-layer CNN (PDA) that maps the feature input to four-class bit-widths is first inserted. Following PDA, it is a quantizer containing two division and addition operators, and finally, the dequantizer in QAC closely follows behind each convolution operator.\par

\subsubsection{Training Details}
The proposed DQMQ is trained on 1 Nvidia GTX 3090TI GPU. For ImageNet dataset, the framework is trained with 120 epochs. For CIFAR-10 and SVHN, DQMQ is converged in 60 epochs. The batch size is 512 in each training and testing process, and the learning rate starts from 0.1 and is uniformly reduced to 0.001. The hyperparameter $\alpha$, which adjusts the training weight between supervised learning optimization and policy optimization, is empirically set as 0.5 in this work. \par

\begin{table}[tbp]
\renewcommand\arraystretch{1.5}
\centering
\caption{Comparison of Top-1 accuracy of DQMQ with popular Fixed-Precision Quantization methods. Results are of ResNet-18 on ImageNet. `FP' means full-precision training, and $\delta_{acc}$ means the accuracy difference between quantization model and full-precision model. `W-Comp' is the weight compression rate. 
}
\scalebox{1}[0.9]{
\begin{tabular}{lccccc}
\bottomrule \hline
                              & \multicolumn{4}{c}{\textbf{Fixed-Precision Quantization}}     &   \\\hline

\textbf{Method}    & \multicolumn{2}{|c}{3-bit}       & \multicolumn{2}{c|}{4-bit}     & \textbf{W-Comp}\\ \hline

                   & \multicolumn{1}{|c}{Acc($\%$)}  & $\delta_{acc}$              & Acc($\%$)       & \multicolumn{1}{c|}{{$\delta_{acc}$} }     & \\ \hline

\multicolumn{1}{c|}{PACT}           & 68.30          & -1.90                       & 69.20            & \multicolumn{1}{c|}{-1.00}                 &   \\        
 
\multicolumn{1}{c|}{LQNet}          & 68.20           & -2.10                       & 69.30           & \multicolumn{1}{c|}{-1.00}                    &    \\           
\multicolumn{1}{c|}{DNAS}           & 68.70          & -2.30                       & 70.60            & \multicolumn{1}{c|}{-0.40}                   & 4.00X/2.00X     \\  
\multicolumn{1}{c|}{US}             & 69.40         & -1.50                    & 70.50               & \multicolumn{1}{c|}{-0.40}                  &   \\ 
 
                        \multicolumn{1}{c|}{SAT}                    & 69.30                         &-1.90   & 70.30                         & \multicolumn{1}{c|}{0.10}                          &                         \\ 
                        \multicolumn{1}{c|}{TQT}                           & *                           & *                             & 69.51                        & \multicolumn{1}{c|}{-0.77 }                          &                        \\ \hline

\multicolumn{1}{c|}{\textbf{DQMQ (ours)}}        &\multicolumn{4}{c|}{\underline{\textbf{71.47}}}     & \underline{\textbf{5.69X}}             \\ \hline
\multicolumn{1}{c|}{FP($\%$)}        &\multicolumn{4}{c|}{70.28}     & 1.00X       \\ \hline
 
\multicolumn{1}{c|}{Size(MB)} & \multicolumn{2}{c}{13.28 v.s. {\underline{\textbf{7.83}}}}    &\multicolumn{2}{c|}{24.42 v.s. {\underline{\textbf{7.83}}}}  &

\\\bottomrule \hline

\end{tabular}
}
\label{table1}
\end{table}

\begin{table}[tbp]
\renewcommand\arraystretch{1.5}
\centering
\caption{Comparison of Top-1 accuracy of DQMQ with other popular  Mixed-Precision Quantization methods. Results are of ResNet-18 on ImageNet. Data with a `*' refers to certain constrained conditions (Details can be found in the references).}
\scalebox{1}[0.9]{
\begin{tabular}{ccccc}
\bottomrule \hline

                              & \multicolumn{3}{c}{\textbf{Mixed-Precision Quantization}}     &   \\\hline
\multicolumn{1}{c}{\textbf{Method}}                     & Acc($\%$)     & $\delta_{acc}$   &Size(MB)        &W-Comp     \\ \hline

          DQ                     & 70.08                           & -0.20      &$7.54^*$     & 5.90X                            \\ 
          AutoQ                 & $68.20^*$                        &$-0.70^*$    &$5.19^*$     & 8.58X                             \\ 
          HAWQ                 & 70.22                            & -1.25      &$8.24^*$   &5.40X                            \\ 
\textbf{DQMQ (ours)}  & \underline{\textbf{71.47}}  & \underline{\textbf{1.19}}  &\underline{\textbf{7.83}} &\underline{\textbf{5.69X}}          \\ 
\multicolumn{1}{c}{FP($\%$)}                 & 70.28                            & 0.00      &$8.24^*$   &1.00X                            \\ 
 
\bottomrule
\end{tabular}
}
\label{table2}
\end{table}

\begin{table}[]
\renewcommand\arraystretch{1.5}
\centering
\caption{Quantization results on CIFAR-10 (Mixed) and SVHN. `W/A' are the weights' and activations' bit-widths respectively. `MP' refers to mixed-precision quantization.}
\scalebox{1}[0.9]{
\begin{tabular}{cccc}
\bottomrule
\textbf{Dataset}            & \textbf{Bits(W/A)} & \textbf{Method}         & \textbf{Acc (Top-1)($\%$)}                                                                                       \\ \bottomrule
                            & 32/32                                        & FP                 & 76.56         \\
CIFAR-10                    & MP                                           & HAWQ               & 81.9       \\
                          & MP                                           & \textbf{DQMQ (ours)}                & \underline{\textbf{82.03}} \\ \hline
                            & 32/32                                        & FP                 & 76.56         \\
CIFAR-10                    & MP                                           & HAWQ               & 69.76        \\
(Mixed)                     & MP                                           & \textbf{DQMQ (ours)}                & \underline{\textbf{81.43}} \\ \hline
                           & 32/32                                        & FP                                           &  *                                                  \\
SVHN                       & MP                                           & HAWQ                                                       & 74.32\\    (Mixed)
                           & MP                                           &\textbf{DQMQ (ours)} & \underline{\textbf{79.02}}  \\ \hline 
                                                       & 32/32                                        & FP                                           &  *                                                  \\
SVHN                        & MP                                           & HAWQ                                                       & 78.13                                             \\
                           & MP                                           &\textbf{DQMQ (ours)} & \underline{\textbf{78.91}}  \\ \bottomrule
\end{tabular}}
\label{table3}
\end{table}

\subsection{Performance Comparison with SOTA} 
In this section, we compare the task performance of our data-aware mixed-precision quantization method with various SOTA methods including both fixed-precision and mixed-precision. More specifically, for fixed-precision quantization, we compare DQMQ with various works including PACT, LQNet, DNAS, US, SAT, TQT networks~\cite{pact,lqnet,dnas,us,sat,tqt}. For mixed-precision quantization works, DQMQ is compared with some of the most representative works including DQ, AutoQ, HAWQ~\cite{dq,AutoQ,hawq}. 
All methods in this section are evaluated on the ImageNet dataset and ResNet-18 architecture, and we use task accuracy and model size reduction as the evaluation measurements for performance comparisons.

Table.\ref{table1} shows the results of the comparison between DQMQ and fixed-precision quantization works. Note that in the last line named `Size', the 13.28M and 24.42M are the model sizes under fixed-precision quantization with 3-bits and 4-bits, respectively, and 7.83 is the size of the quantized model from DQMQ with mixed bit-widths.
It can be clearly seen that for fixed-precision quantization, nearly all methods suffer from the accuracy drop caused by model quantization except SAT, which has a small accuracy gain of $0.1\%$ on 4-bit quantization. Conversely, DQMQ shows greater accuracy gains with $1.19\%$ accuracy improvement.
What's more, for fixed methods, the compression rate of model size is limited by the integer bit-width division. Excluding binary and tertiary quantization that bring a large drop in accuracy, the highest model compression rate in most fixed-precision methods does not exceed 4 times.
%
The comparison results with mixed-precision are shown in Table.\ref{table2}.
Similar to the static quantization result, nearly all existing mixed methods show accuracy degradation after quantization except AutoQ, which has a small accuracy benefit of $0.7\%$. 
Regarding model quantization as a perturbation to the original model parameters, the drop in accuracy in most previous works reveals their poor perturbation adversarial.
Conversely, the reward scheme in data quality-aware quantization helps learn the optimal quantitative policy when facing diverse samples, greatly reducing the parameters entropy and improving the quantization robustness~\cite{lottery}. 
Finally, we compare the task performance of HAWQ and DQMQ based on CIFAR-10 (Mixed) and SVHN (Mixed) with ResNet-18 architecture. We test both the quantized models on various data qualities to demonstrate if the proposed DQMQ has the ability to adapt itself to different test distributions. The results are listed in Table.~\ref{table3}. In CIFAR-10 (Mixed) and SVHN (Mixed), the data quality ranges from the worst Gaussian blur to better image sharpening widely. It is obvious that our DQMQ can achieve higher test accuracy than HAWQ on all dataset. To be specific, on the original CIFAR-10 and SVHN with uniform data quality, as predicted, due to the lack of rich data quality knowledge in training stage, the performance of DQMQ and HAWQ is almost equal. However, in data domains where the data distribution changes drastically (mixed datasets), DQMQ can make better quantization decisions and reap higher accuracy and robustness than HAWQ. 

\begin{figure}[ht]
	\centering
	\subfloat[Blurred data]{
		\includegraphics[width =0.8\columnwidth]{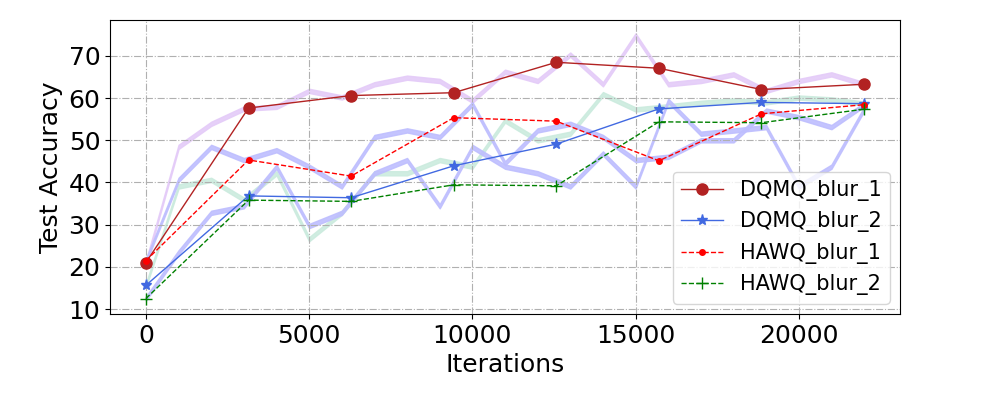} 
		
	}\\
	\subfloat[Standard data]{
		\includegraphics[width =0.8\columnwidth]{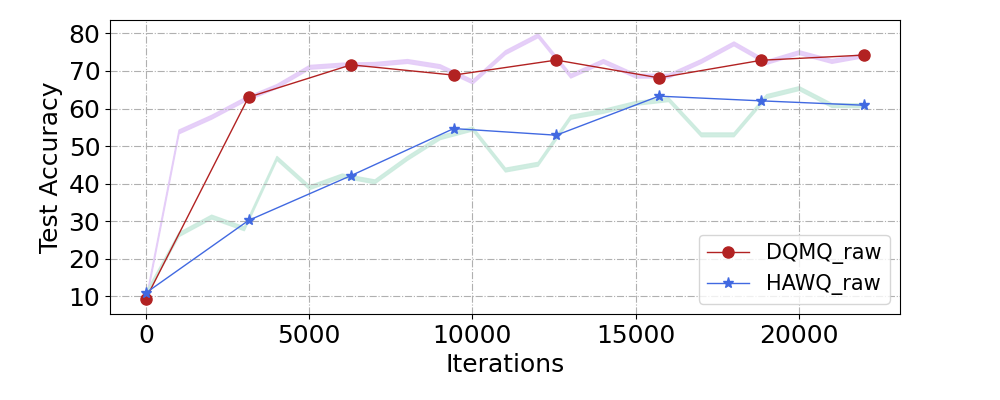} 
	}\\
	\subfloat[Sharpened data]{
		\includegraphics[width =0.8\columnwidth]{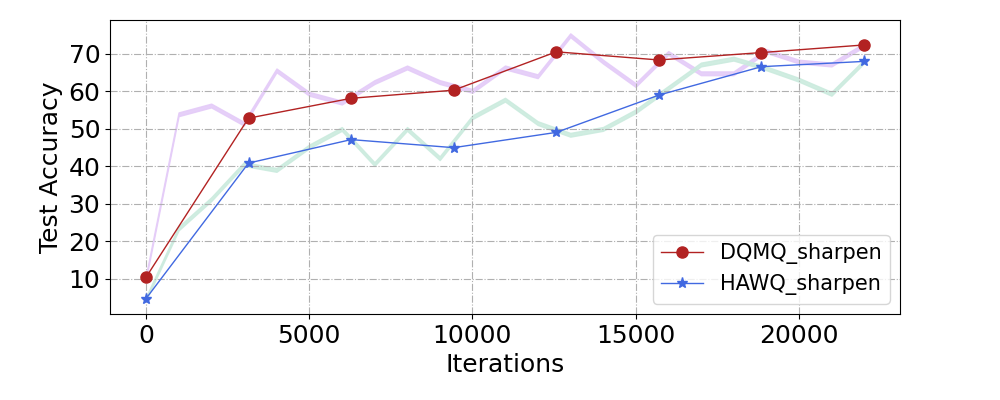} 
	}
	\caption{Comparison of accuracy (Top-1)($\%$) of DQMQ and HAWQ of ResNet-18 on CIFAR-10 with different image qualities.}
	\label{fig3}
\end{figure}

\subsection{Correlation Exploration Between Data Quality and Quantization Sensitivity} 
In this subsection, we would prove the strong correlation between data quality with layer quantization sensitivity to emphasize the strong necessity of our research. 
The underlying motivation origins from the fact that mobile devices are most likely working in edge environments with dynamically changing contexts. The uncontrollable data qualities in running time could have a large distribution deviation from that of training time and cause the vanilla quantization to suffer significant performance degradation.
In order to find a more robust model quantization in the face of changing data quality, it is important to explore the relationship between data quality and optimal quantization bit-width.
 
\begin{figure}[t]
	\centering

	\subfloat[ResNet-18: Layer sensitivity under different data qualities.]{
		\includegraphics[width =0.92\columnwidth]{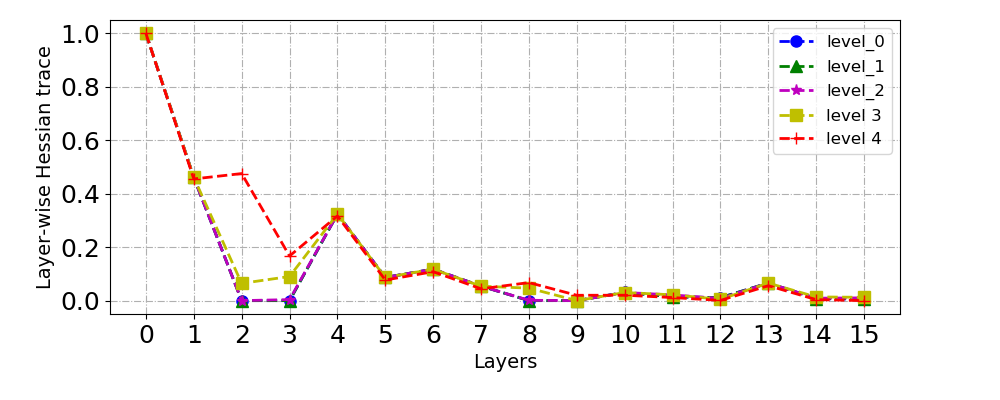} 
	}\\
	\subfloat[ResNet-18: Enlarged view of the last eight layers.]{
		\includegraphics[width =0.8\columnwidth]{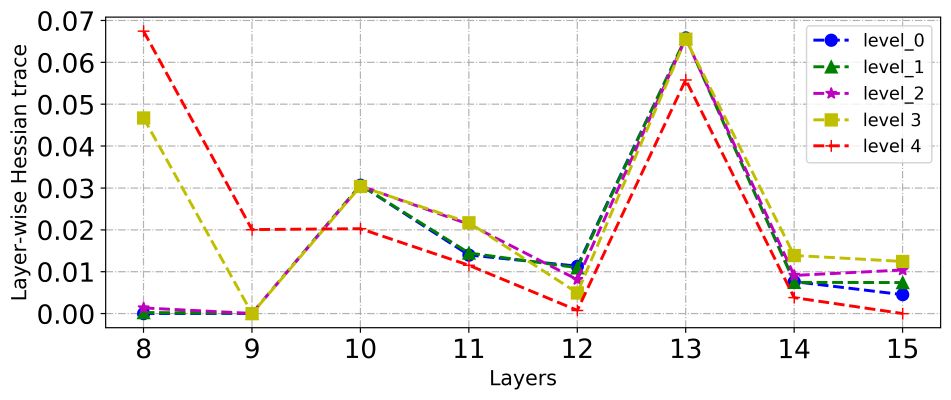} 
	}\\
	\subfloat[ResNet-20: Layer sensitivity under different data qualities.]{
		\includegraphics[width =0.92\columnwidth]{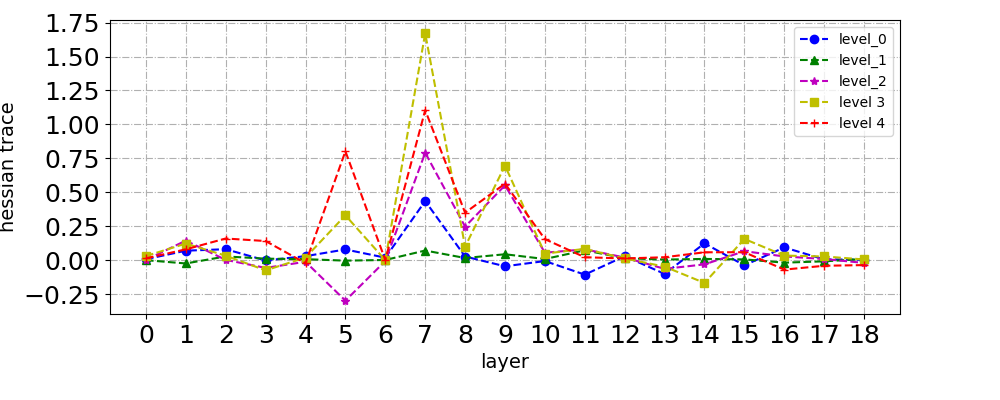} 
	}
	\caption{Comparison of the layer-wise quantization sensitivities of DQMQ on ResNet-18 under CIFAR-10 with five different data qualities.} 
	\label{hessianinfo}
\end{figure}

\begin{figure}[t]
\centering
\includegraphics[width =0.92\columnwidth]{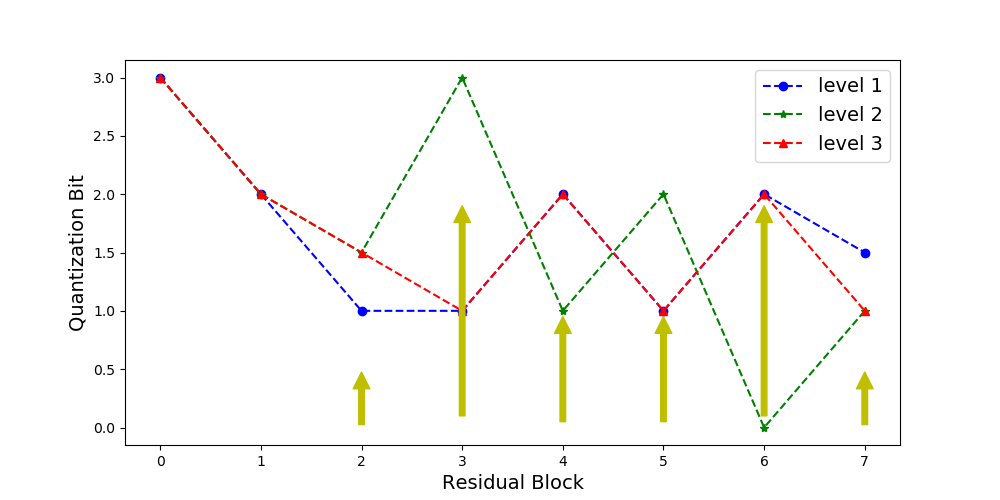}
\caption{The quantization precision decisions made by DQMQ under different data qualities and ResNet-18. The horizontal axis is the index of the convolutional network in the network, and the vertical axis is the quantization accuracy of this layer determined by DQMQ. Yellow arrows represent the maximum difference of quantization bit-widths for each layer under three different data qualities.}
\label{fig4}
\end{figure}

\begin{table}[]
\renewcommand\arraystretch{1.5}
\centering
\caption{Comparison of Top-1 accuracy between mixed-precision quantization methods HAWQ and DQMQ under three data qualities.Results are of ResNet-18 on specially processed CIFAR-10.}
\scalebox{1}[0.9]{
\begin{tabular}{cccc}
\bottomrule
\textbf{Method}          & \textbf{Data Quality} & \textbf{Acc (Top-1)(\%)} \\   \bottomrule
                         & blurring        & 67.19         \\ 
HAWQ          & standard     & 81.90           \\ 
                         & sharpening        & 80.31         \\ \hline
                         & blurring      & \underline{\textbf{76.50}}         \\ 
\textbf{DQMQ (ours)}       & standard     & \underline{\textbf{82.03}}         \\ 
                         & sharpening      & \underline{\textbf{82.34}}         \\ \bottomrule
\end{tabular}}
\label{table4}
\end{table}
Thus, extensive experiments are conduct on datasets with mixed data qualities and the results are shown in Fig.~\ref{hessianinfo} and Fig.~\ref{fig4}. 
As shown in Fig.~\ref{hessianinfo}, since the trace of the Hessian matrix is positively correlated with the relative quantization sensitivities between layers~\cite{HAWQ-v2}, we compare layers' hessian trace under five data qualities to observe the changes in the optimal quantization precision for each layer on both ResNet-18 and ResNet-20. 
In ResNet-18, as data quality gets worse, sensitivities of deeper layers begin to increase, while the layer-wise sensitivity differences show a downward trend.   
In ResNet-20, the results show that with increasing image noise, the degree of the turbulence of hessian curve gradually increases, which confirms the fact that as data quality changing, layer-wise relative quantization sensitivity should fluctuate dynamically rather than remain static as existing methods. Data quality would be a unavoidable factor in quantization settings searching.
What's more, as shown in Fig.~\ref{fig4}, the yellow arrows represent the maximum quantization bit-width difference for each layer under three data qualities. It is clearly that the layers in the middle and closer to the classifier are more sensitive to the changing of data quality. These layers make totally different quantization bit-width decisions when data quality changes. 
At this point, we can claim that: \textit{under ambiguous test distributions about data quality, DQMQ could dynamically make the model quantization adapt to the data variation and thus lead to better robustness at diverse data qualities}.

To be more precise, we also compare the accuracy of the proposed DQMQ with HAWQ under three different data qualities. Both methods are conducted on ResNet-18 and CIFAR-10 (Mixed). Results are shown in Table.~\ref{table4}. For HAWQ, when data quality increases from standard quality(unprocessed CIFAR-10) to better quality(Sharpening CIFAR-10), 
contrary to expectations, there is a slight drop in model accuracy even with better data quality.
The disordered relationship between the accuracy and input data qualities of HAWQ 
illustrates that for the same bit-width, the performance of the quantized model is not simply proportional to the data quality. In the relationship chain of "data quality $\rightarrow$ quantization settings $\rightarrow$ quantization performance", the first and last items are not directly related and lost the expected consistency. 
HAWQ takes only model information into consideration when make quantization decisions and ignores changes in data. Different from that, DQMQ uses both model  and data information to dynamically make the quantization bit-width decision. DQMQ not only shows a higher prediction accuracy than SOTA under the same quality, but also keeps showing an accuracy ascension as data quality gets better. DQMQ restores the positive relationship between model performance and data quality in quantitative scenarios, which also demonstrates its data quality-adaptive ability from one-shot quantization training.

\section{Conclusion}
In this paper, we proposed a novel data-quality aware model quantization method (DQMQ) to achieve more robust compressed models in the edge environment. 
Our method is the first attempt that focuses on the perspective of data rather than the model architecture or algorithm, to improve both the generalization and robustness of quantized models. 
Our method can dynamically adapt the quantization precision to different data qualities with layer-wise multiple bit-widths synchronously with task performance. Moreover, we also incorporate the decoupled bit-width decision-making and quantization training into a much more efficient and coherent one-shot training scheme. 
Extensive experiments verified the effectiveness of the proposed method.

\ifCLASSOPTIONcaptionsoff
  \newpage
\fi
\bibliographystyle{IEEEtran}
\bibliography{bibfile}

\begin{thebibliography}{10}
\providecommand{\url}[1]{#1}
\csname url@samestyle\endcsname
\providecommand{\newblock}{\relax}
\providecommand{\bibinfo}[2]{#2}
\providecommand{\BIBentrySTDinterwordspacing}{\spaceskip=0pt\relax}
\providecommand{\BIBentryALTinterwordstretchfactor}{4}
\providecommand{\BIBentryALTinterwordspacing}{\spaceskip=\fontdimen2\font plus
\BIBentryALTinterwordstretchfactor\fontdimen3\font minus
  \fontdimen4\font\relax}
\providecommand{\BIBforeignlanguage}[2]{{%
\expandafter\ifx\csname l@#1\endcsname\relax
\typeout{** WARNING: IEEEtran.bst: No hyphenation pattern has been}%
\typeout{** loaded for the language `#1'. Using the pattern for}%
\typeout{** the default language instead.}%
\else
\language=\csname l@#1\endcsname
\fi
#2}}
\providecommand{\BIBdecl}{\relax}
\BIBdecl

\bibitem{Fiorio15}
C.~Fiorio, \emph{{a}lgorithm2e.sty---package for algorithms}, Oct. 2015,
  \url{http://www.ctan.org/pkg/algorithm2e}.

\bibitem{post}
X.~Liu, M.~Ye, D.~Zhou, and Q.~Liu, ``Post-training quantization with multiple
  points: Mixed precision,'' in \emph{Thirty-Fifth {AAAI} Conference on
  Artificial Intelligence, {AAAI} 2021, Thirty-Third Conference on Innovative
  Applications of Artificial Intelligence, {IAAI} 2021, The Eleventh Symposium
  on Educational Advances in Artificial Intelligence, {EAAI} 2021, Virtual
  Event, February 2-9, 2021}.\hskip 1em plus 0.5em minus 0.4em\relax {AAAI}
  Press, 2021, pp. 8697--8705.

\bibitem{HAWQ-v2}
Z.~Dong, Z.~Yao, D.~Arfeen, A.~Gholami, M.~W. Mahoney, and K.~Keutzer,
  ``{HAWQ-V2:} hessian aware trace-weighted quantization of neural networks,''
  in \emph{Advances in Neural Information Processing Systems 33: Annual
  Conference on Neural Information Processing Systems 2020, NeurIPS 2020,
  December 6-12, 2020, virtual}, 2020.

\bibitem{deep}
S.~Han, H.~Mao, and W.~J. Dally, ``Deep compression: Compressing deep neural
  network with pruning, trained quantization and huffman coding,'' in \emph{4th
  International Conference on Learning Representations, {ICLR} 2016, San Juan,
  Puerto Rico, May 2-4, 2016, Conference Track Proceedings}, 2016.

\bibitem{sat}
Q.~Jin, L.~Yang, and Z.~Liao, ``Towards efficient training for neural network
  quantization,'' \emph{CoRR}, vol. abs/1912.10207, 2019.

\bibitem{irnet}
H.~Qin, R.~Gong, X.~Liu, M.~Shen, Z.~Wei, F.~Yu, and J.~Song, ``Forward and
  backward information retention for accurate binary neural networks,'' in
  \emph{2020 {IEEE/CVF} Conference on Computer Vision and Pattern Recognition,
  CVPR 2020}.\hskip 1em plus 0.5em minus 0.4em\relax Computer Vision Foundation
  / {IEEE}, 2020, pp. 2247--2256.

\bibitem{hawq}
Z.~Dong, Z.~Yao, A.~Gholami, M.~W. Mahoney, and K.~Keutzer, ``{HAWQ:} hessian
  aware quantization of neural networks with mixed-precision,'' in \emph{2019
  {IEEE/CVF} International Conference on Computer Vision, {ICCV} 2019}.\hskip
  1em plus 0.5em minus 0.4em\relax {IEEE}, 2019, pp. 293--302.

\bibitem{HAWQ-V3}
Z.~Yao, Z.~Dong, and Z.~Zheng, ``{HAWQ-V3:} dyadic neural network
  quantization,'' in \emph{Proceedings of the 38th International Conference on
  Machine Learning, {ICML} 2021, 18-24 July 2021, Virtual Event}, ser.
  Proceedings of Machine Learning Research, vol. 139.\hskip 1em plus 0.5em
  minus 0.4em\relax {PMLR}, 2021, pp. 11\,875--11\,886.

\bibitem{daam}
S.~Zhao, T.~Yue, and X.~Hu, ``Distribution-aware adaptive multi-bit
  quantization,'' in \emph{{IEEE} Conference on Computer Vision and Pattern
  Recognition, {CVPR} 2021, virtual, June 19-25, 2021}, 2021, pp. 9281--9290.

\bibitem{FracBits}
L.~Yang and Q.~Jin, ``Fracbits: Mixed precision quantization via fractional
  bit-widths,'' in \emph{Thirty-Fifth {AAAI} Conference on Artificial
  Intelligence, {AAAI} 2021}.\hskip 1em plus 0.5em minus 0.4em\relax {AAAI}
  Press, 2021, pp. 10\,612--10\,620.

\bibitem{DBLP:journals/corr/abs-1806-08342}
R.~Krishnamoorthi, ``Quantizing deep convolutional networks for efficient
  inference: {A} whitepaper,'' \emph{CoRR}, vol. abs/1806.08342, 2018.

\bibitem{DBLP:conf/cvpr/JacobKCZTHAK18}
B.~Jacob, S.~Kligys, and B.~Cheno, ``Quantization and training of neural
  networks for efficient integer-arithmetic-only inference,'' in \emph{2018
  {IEEE} Conference on Computer Vision and Pattern Recognition, {CVPR} 2018,
  Salt Lake City, UT, USA, June 18-22, 2018}, 2018, pp. 2704--2713.

\bibitem{DBLP:conf/icmcs/ZhuZL18}
X.~Zhu, W.~Zhou, and H.~Li, ``Adaptive layerwise quantization for deep neural
  network compression,'' in \emph{2018 {IEEE} International Conference on
  Multimedia and Expo, {ICME} 2018}, 2018, pp. 1--6.

\bibitem{haq}
K.~Wang, Z.~Liu, Y.~Lin, J.~Lin, and S.~Han, ``{HAQ:} hardware-aware automated
  quantization with mixed precision,'' in \emph{{IEEE} Conference on Computer
  Vision and Pattern Recognition, CVPR 2019}, 2019, pp. 8612--8620.

\bibitem{dorefa}
S.~Zhou, Z.~Ni, X.~Zhou, H.~Wen, Y.~Wu, and Y.~Zou, ``Dorefa-net: Training low
  bitwidth convolutional neural networks with low bitwidth gradients,''
  \emph{CoRR}, vol. abs/1606.06160, 2016.

\bibitem{DBLP:journals/nn/YangDWYXL20}
Y.~Yang, L.~Deng, S.~Wu, T.~Yan, Y.~Xie, and G.~Li, ``Training high-performance
  and large-scale deep neural networks with full 8-bit integers,'' \emph{Neural
  Networks}, vol. 125, pp. 70--82, 2020.

\bibitem{tint}
F.~Zhu, R.~Gong, F.~Yu, X.~Liu, Y.~Wang, Z.~Li, X.~Yang, and J.~Yan, ``Towards
  unified int8 training for convolutional neural network,'' in \emph{2020
  {IEEE/CVF} Conference on Computer Vision and Pattern Recognition, CVPR 2020},
  2020, pp. 1966--1976.

\bibitem{DBLP:journals/corr/abs-1812-00090}
B.~Wu, Y.~Wang, P.~Zhang, Y.~Tian, P.~Vajda, and K.~Keutzer, ``Mixed precision
  quantization of convnets via differentiable neural architecture search,''
  \emph{CoRR}, vol. abs/1812.00090, 2018.

\bibitem{pact}
J.~Choi, Z.~Wang, S.~Venkataramani, P.~I. Chuang, V.~Srinivasan, and
  K.~Gopalakrishnan, ``Pact: Parameterized clipping activation for quantized
  neural networks,'' \emph{CoRR}, vol. abs/1805.06085, 2018.

\bibitem{AutoQ}
Q.~Lou, F.~Guo, M.~Kim, L.~Liu, and L.~Jiang, ``Autoq: Automated kernel-wise
  neural network quantization,'' in \emph{8th International Conference on
  Learning Representations, {ICLR} 2020, Addis Ababa, Ethiopia, April 26-30,
  2020}, 2020.

\bibitem{DBLP:conf/iclr/WuLCS18}
S.~Wu, G.~Li, F.~Chen, and L.~Shi, ``Training and inference with integers in
  deep neural networks,'' in \emph{6th International Conference on Learning
  Representations, {ICLR} 2018, Vancouver, BC, Canada, April 30 - May 3, 2018,
  Conference Track Proceedings}, 2018.

\bibitem{DBLP:conf/nips/WangCBCG18}
N.~Wang, J.~Choi, D.~Brand, C.~Chen, and K.~Gopalakrishnan, ``Training deep
  neural networks with 8-bit floating point numbers,'' in \emph{Advances in
  Neural Information Processing Systems 31: Annual Conference on Neural
  Information Processing Systems 2018, NeurIPS 2018}, 2018, pp. 7686--7695.

\bibitem{DBLP:conf/iclr/0002MMKAB0VKGHD18}
D.~Das, N.~Mellempudi, D.~Mudigere, and D.~D. Kalamkar, ``Mixed precision
  training of convolutional neural networks using integer operations,'' in
  \emph{6th International Conference on Learning Representations, {ICLR} 2018,
  Vancouver, BC, Canada, April 30 - May 3, 2018, Conference Track Proceedings},
  2018.

\bibitem{DBLP:conf/nips/KosterWWNBCEHHK17}
U.~K{\"{o}}ster, T.~Webb, X.~Wang, and M.~Nassa, ``Flexpoint: An adaptive
  numerical format for efficient training of deep neural networks,'' in
  \emph{Advances in Neural Information Processing Systems 30: Annual Conference
  on Neural Information Processing Systems 2017, December 4-9, 2017, Long
  Beach, CA, {USA}}, 2017, pp. 1742--1752.

\bibitem{skip}
X.~Wang, F.~Yu, Z.~Dou, T.~Darrell, and J.~E. Gonzalez, ``Skipnet: Learning
  dynamic routing in convolutional networks,'' in \emph{Computer Vision -
  {ECCV} 2018 - 15th European Conference, Munich, Germany, September 8-14,
  2018, Proceedings, Part {XIII}}, vol. 11217, 2018, pp. 420--436.

\bibitem{ada}
Q.~Jin, L.~Yang, and Z.~Liao, ``Adabits: Neural network quantization with
  adaptive bit-widths,'' in \emph{2020 {IEEE/CVF} Conference on Computer Vision
  and Pattern Recognition, {CVPR} 2020, Seattle, WA, USA, June 13-19, 2020},
  2020, pp. 2143--2153.

\bibitem{lqnet}
D.~Zhang, J.~Yang, D.~Ye, and G.~Hua, ``Lq-nets: Learned quantization for
  highly accurate and compact deep neural networks,'' in \emph{Computer Vision
  - {ECCV} 2018 - 15th European Conference, Munich, Germany, September 8-14,
  2018, Proceedings, Part {VIII}}, ser. Lecture Notes in Computer Science, vol.
  11212, 2018, pp. 373--390.

\bibitem{dnas}
B.~Wu, X.~Dai, and P.~Zhang, ``Fbnet: Hardware-aware efficient convnet design
  via differentiable neural architecture search,'' in \emph{{IEEE} Conference
  on Computer Vision and Pattern Recognition, {CVPR} 2019, Long Beach, CA, USA,
  June 16-20, 2019}, 2019, pp. 10\,734--10\,742.

\bibitem{us}
Z.~Guo, X.~Zhang, H.~Mu, W.~Heng, Z.~Liu, Y.~Wei, and J.~Sun, ``Single path
  one-shot neural architecture search with uniform sampling,'' in
  \emph{Computer Vision - {ECCV} 2020 - 16th European Conference, Glasgow, UK,
  August 23-28, 2020, Proceedings, Part {XVI}}, ser. Lecture Notes in Computer
  Science, vol. 12361, 2020, pp. 544--560.

\bibitem{tqt}
S.~R. Jain, A.~Gural, M.~Wu, and C.~Dick, ``Trained uniform quantization for
  accurate and efficient neural network inference on fixed-point hardware,''
  \emph{CoRR}, vol. abs/1903.08066, 2019.

\bibitem{dq}
S.~Uhlich, L.~Mauch, and F.~Cardinaux, ``Mixed precision dnns: All you need is
  a good parametrization,'' in \emph{8th International Conference on Learning
  Representations, {ICLR} 2020, Addis Ababa, Ethiopia, April 26-30,
  2020}.\hskip 1em plus 0.5em minus 0.4em\relax OpenReview.net, 2020.

\bibitem{lottery}
J.~Frankle and M.~Carbin, ``The lottery ticket hypothesis: Finding sparse,
  trainable neural networks,'' in \emph{7th International Conference on
  Learning Representations, {ICLR} 2019, New Orleans, LA, USA, May 6-9, 2019},
  2019.

\bibitem{bnn}
F.~Zheng, C.~Deng, and H.~Huang, ``Binarized neural networks for
  resource-efficient hashing with minimizing quantization loss,'' in
  \emph{Proceedings of the Twenty-Eighth International Joint Conference on
  Artificial Intelligence, {IJCAI} 2019, Macao, China, August 10-16, 2019},
  S.~Kraus, Ed.\hskip 1em plus 0.5em minus 0.4em\relax ijcai.org, 2019, pp.
  1032--1040.

\bibitem{tnn}
C.~Zhu, S.~Han, H.~Mao, and W.~J. Dally, ``Trained ternary quantization,'' in
  \emph{5th International Conference on Learning Representations, {ICLR} 2017,
  Toulon, France, April 24-26, 2017, Conference Track Proceedings}.\hskip 1em
  plus 0.5em minus 0.4em\relax OpenReview.net, 2017.

\bibitem{int8}
K.~Zhao, S.~Huang, P.~Pan, Y.~Li, Y.~Zhang, Z.~Gu, and Y.~Xu, ``Distribution
  adaptive {INT8} quantization for training cnns,'' in \emph{Thirty-Fifth
  {AAAI} Conference on Artificial Intelligence, {AAAI} 2021, Thirty-Third
  Conference on Innovative Applications of Artificial Intelligence, {IAAI}
  2021, The Eleventh Symposium on Educational Advances in Artificial
  Intelligence, {EAAI} 2021, Virtual Event, February 2-9, 2021}.\hskip 1em plus
  0.5em minus 0.4em\relax {AAAI} Press, 2021, pp. 3483--3491.

\bibitem{qat}
\BIBentryALTinterwordspacing
B.~Jacob, S.~Kligys, B.~Chen, M.~Zhu, M.~Tang, A.~G. Howard, H.~Adam, and
  D.~Kalenichenko, ``Quantization and training of neural networks for efficient
  integer-arithmetic-only inference,'' in \emph{2018 {IEEE} Conference on
  Computer Vision and Pattern Recognition, {CVPR} 2018, Salt Lake City, UT,
  USA, June 18-22, 2018}.\hskip 1em plus 0.5em minus 0.4em\relax Computer
  Vision Foundation / {IEEE} Computer Society, 2018, pp. 2704--2713. [Online].
  Available:
  \url{http://openaccess.thecvf.com/content\_cvpr\_2018/html/Jacob\_Quantization\_and\_Training\_CVPR\_2018\_paper.html}
\BIBentrySTDinterwordspacing

\bibitem{Lu2019Lip}
L.~Lu, J.~Yu, Y.~Chen, H.~Liu, Y.~Zhu, L.~Kong, and M.~Li, ``Lip reading-based
  user authentication through acoustic sensing on smartphones,'' vol.~27,
  no.~1, 2019, pp. 1--14.

\bibitem{wang2022survey}
Y.~Wang, J.~Wang, W.~Zhang, Y.~Zhan, S.~Guo, Q.~Zheng, and X.~Wang, ``A survey
  on deploying mobile deep learning applications: A systemic and technical
  perspective,'' \emph{Digital Communications and Networks}, vol.~8, no.~1, pp.
  1--17, 2022.

\bibitem{zhang2018deepvoice}
H.~Zhang, A.~Wang, D.~Li, and W.~Xu, ``Deepvoice: a voiceprint-based mobile
  health framework for parkinson's disease identification,'' in \emph{2018 IEEE
  EMBS International Conference on Biomedical \& Health Informatics
  (BHI)}.\hskip 1em plus 0.5em minus 0.4em\relax IEEE, 2018, pp. 214--217.

\bibitem{PDVoice}
------, ``Deepvoice: {A} voiceprint-based mobile health framework for
  parkinson's disease identification,'' in \emph{2018 {IEEE} {EMBS}
  International Conference on Biomedical {\&} Health Informatics, {BHI} 2018,
  Las Vegas, NV, USA, March 4-7, 2018}, 2018, pp. 214--217.

\bibitem{Jacob_2018_CVPR}
B.~Jacob, S.~Kligys, B.~Chen, M.~Zhu, M.~Tang, A.~Howard, H.~Adam, and
  D.~Kalenichenko, ``Quantization and training of neural networks for efficient
  integer-arithmetic-only inference,'' in \emph{Proceedings of the IEEE
  Conference on Computer Vision and Pattern Recognition (CVPR)}, June 2018.

\bibitem{ste1}
\BIBentryALTinterwordspacing
Y.~Bengio, N.~L{\'{e}}onard, and A.~C. Courville, ``Estimating or propagating
  gradients through stochastic neurons for conditional computation,''
  \emph{CoRR}, vol. abs/1308.3432, 2013. [Online]. Available:
  \url{http://arxiv.org/abs/1308.3432}
\BIBentrySTDinterwordspacing

\bibitem{ste2}
R.~Gong, X.~Liu, S.~Jiang, T.~Li, P.~Hu, J.~Lin, F.~Yu, and J.~Yan,
  ``Differentiable soft quantization: Bridging full-precision and low-bit
  neural networks,'' in \emph{2019 {IEEE/CVF} International Conference on
  Computer Vision, {ICCV} 2019, Seoul, Korea (South), October 27 - November 2,
  2019}.\hskip 1em plus 0.5em minus 0.4em\relax {IEEE}, 2019, pp. 4851--4860.

\bibitem{esser}
S.~K. Esser, J.~L. McKinstry, D.~Bablani, R.~Appuswamy, and D.~S. Modha,
  ``Learned step size quantization,'' in \emph{8th International Conference on
  Learning Representations, {ICLR} 2020, Addis Ababa, Ethiopia, April 26-30,
  2020}.\hskip 1em plus 0.5em minus 0.4em\relax OpenReview.net, 2020.

\bibitem{uhl}
S.~Uhlich, L.~Mauch, F.~Cardinaux, K.~Yoshiyama, J.~A. Garc{\'{\i}}a,
  S.~Tiedemann, T.~Kemp, and A.~Nakamura, ``Mixed precision dnns: All you need
  is a good parametrization,'' in \emph{8th International Conference on
  Learning Representations, {ICLR} 2020, Addis Ababa, Ethiopia, April 26-30,
  2020}.\hskip 1em plus 0.5em minus 0.4em\relax OpenReview.net, 2020.

\bibitem{updown}
M.~Nagel, R.~A. Amjad, M.~van Baalen, C.~Louizos, and T.~Blankevoort, ``Up or
  down? adaptive rounding for post-training quantization,'' in
  \emph{Proceedings of the 37th International Conference on Machine Learning,
  {ICML} 2020, 13-18 July 2020, Virtual Event}, ser. Proceedings of Machine
  Learning Research, vol. 119.\hskip 1em plus 0.5em minus 0.4em\relax {PMLR},
  2020, pp. 7197--7206.

\bibitem{1}
M.~Nagel, M.~van Baalen, T.~Blankevoort, and M.~Welling, ``Data-free
  quantization through weight equalization and bias correction,'' in \emph{2019
  {IEEE/CVF} International Conference on Computer Vision, {ICCV} 2019, Seoul,
  Korea (South), October 27 - November 2, 2019}.\hskip 1em plus 0.5em minus
  0.4em\relax {IEEE}, 2019, pp. 1325--1334.

\bibitem{2}
R.~Banner, Y.~Nahshan, and D.~Soudry, ``Post training 4-bit quantization of
  convolutional networks for rapid-deployment,'' in \emph{Advances in Neural
  Information Processing Systems 32: Annual Conference on Neural Information
  Processing Systems 2019, NeurIPS 2019, December 8-14, 2019, Vancouver, BC,
  Canada}, 2019, pp. 7948--7956.

\bibitem{fang2020post}
J.~Fang, A.~Shafiee, H.~Abdel-Aziz, D.~Thorsley, G.~Georgiadis, and J.~H.
  Hassoun, ``Post-training piecewise linear quantization for deep neural
  networks,'' in \emph{European Conference on Computer Vision}.\hskip 1em plus
  0.5em minus 0.4em\relax Springer, 2020, pp. 69--86.

\bibitem{p3}
J.~Fang, A.~Shafiee, H.~Abdel{-}Aziz, D.~Thorsley, and G.~Georgiadis,
  ``Post-training piecewise linear quantization for deep neural networks,'' in
  \emph{Computer Vision - {ECCV} 2020 - 16th European Conference, Glasgow, UK,
  August 23-28, 2020, Proceedings, Part {II}}, ser. Lecture Notes in Computer
  Science, vol. 12347.\hskip 1em plus 0.5em minus 0.4em\relax Springer, 2020,
  pp. 69--86.

\bibitem{zeroq}
Y.~Cai, Z.~Yao, Z.~Dong, A.~Gholami, M.~W. Mahoney, and K.~Keutzer, ``Zeroq:
  {A} novel zero shot quantization framework,'' in \emph{2020 {IEEE/CVF}
  Conference on Computer Vision and Pattern Recognition, {CVPR} 2020, Seattle,
  WA, USA, June 13-19, 2020}.\hskip 1em plus 0.5em minus 0.4em\relax Computer
  Vision Foundation / {IEEE}, 2020, pp. 13\,166--13\,175.

\bibitem{tensorrt}
Nvidia, ``Nvidia tensorrt,'' https://developer.nvidia.com/tensorrt, 2018, [Open
  resource on December-2018].

\end{thebibliography}
\end{document}